\documentclass{article}

\PassOptionsToPackage{numbers}{natbib}

\usepackage[preprint]{neurips_2023}

\usepackage{subfigure}
\usepackage{kotex}
\usepackage{multirow}
\usepackage{graphicx}
\usepackage{amsmath}
\usepackage{algorithm}
\usepackage{algpseudocode}
\usepackage{wrapfig}
\usepackage{comment}        

\usepackage[utf8]{inputenc} 
\usepackage[T1]{fontenc}    
\usepackage{hyperref}       
\usepackage{url}            
\usepackage{booktabs}       
\usepackage{amsfonts}       
\usepackage{nicefrac}       
\usepackage{microtype}      
\usepackage{xcolor}         

\newcommand{\our}{Diffusion-Stego}
\newcommand{\zm}{message noise} 

\title{\our{}: Training-free Diffusion\\ Generative Steganography via Message Projection }

\author{%
 Daegyu Kim~~~~~Chaehun Shin~~~~~Jooyoung Choi~~~~~Dahuin Jung~~~~~Sungroh Yoon\thanks{Corresponding Author}\\
 Data Science and AI Laboratory, ECE, Seoul National University\\
}

\begin{document}

\maketitle



\begin{abstract}

Generative steganography is the process of hiding secret messages in generated images instead of cover images.
Existing studies on generative steganography use GAN or Flow models to obtain high hiding message capacity and anti-detection ability over cover images. 
However, they create relatively unrealistic stego images because of the inherent limitations of generative models.
We propose \our{}, a generative steganography approach based on diffusion models which outperform other generative models in image generation.
\our{} projects secret messages into latent noise of diffusion models and generates stego images with an iterative denoising process.
Since the naive hiding of secret messages into noise boosts visual degradation and decreases extracted message accuracy, we introduce message projection, which hides messages into noise space while addressing these issues.
We suggest three options for message projection to adjust the trade-off between extracted message accuracy, anti-detection ability, and image quality.
\our{} is a training-free approach, so we can apply it to pre-trained diffusion models which generate high-quality images, or even large-scale text-to-image models, such as Stable diffusion.
\our{} achieved a high capacity of messages (3.0 bpp of binary messages with 98\% accuracy, and 6.0 bpp with 90\% accuracy) as well as high quality (with a FID score of 2.77 for 1.0 bpp on the FFHQ 64$\times$64 dataset) that makes it challenging to distinguish from real images in the PNG format.

\end{abstract}

\section{Introduction}

Image steganography is the process that aims at hiding secret messages in images so that the secret messages are not detected or exposed by third-party players. 
Traditional image steganography methods~\cite{morkel2005overview,johnson1998exploring} conceal secret messages within a natural cover image.
The sender transmits the cover image containing the secret messages, termed a stego image, to the receiver, who extracts the hidden messages from the stego image.
On the contrary, the third-party players attempt to discriminate the stego images by training steganalyzer models~\cite{xu2016structural,ye2017deep,fridrich2012rich}, which classify over the cover images and stego images.

Generative steganography methods~\cite{wu2014steganography,hu2018novel} have been proposed to deceive steganalyzer models. 
These approaches apply deep generative models that synthesize stego images from secret messages without using cover images.
It makes them less vulnerable to steganalyzer models, as there are no cover images for the steganalyzer to train on.
Recent generative steganography studies~\cite{wei2022generative,zhou2022secret,wei2022generative2} using Generative Adversarial Networks (GAN)~\cite{goodfellow2020generative} or Flow~\cite{kingma2018glow} models as a generator have been proposed.
In contrast to their anti-detection ability and high hiding capacity, they relatively lack image fidelity due to the limitations of the generative models they use.

Therefore, we explore utilizing diffusion-based generative models for steganography.
Diffusion models~\cite{sohl2015deep, ho2020denoising} are recently popular generative models, which generate high-quality images~\cite{ramesh2022hierarchical,saharia2022photorealistic,rombach2022high} with an iterative sampling process.
Recent studies~\cite{karras2022elucidating,xu2023pfgm++} have utilized diffusion models and deterministic samplers~\cite{song2020denoising,song2020score,karras2022elucidating} for generating high quality images.

In this paper, we propose \our{}, a powerful generative steganography approach that utilizes diffusion models and deterministic sampler.
As illustrated in Figure~\ref{fig:ov}(a), \our{} hides secret messages in noise and generates stego images using the deterministic sampler without re-training the diffusion models.
Due to the invertible property of the deterministic sampler, \our{} can generate and extract messages using a single diffusion model.
This allows the sender and the receiver to communicate secret messages while sharing only a diffusion model and a hiding method.

However, we have identified that there are two challenges to utilizing a diffusion model in steganography.
First, diffusion models cannot generate images from noise replaced with binary messages, which is the naive approach of hiding messages in the noise.
Second, the accumulation of slight errors during the reverse process of the deterministic sampler leads to a drop in the extracted message accuracy.
To address these challenges, we propose a novel technique called \textit{message projection}.
Message projection modifies noise to the extent that it does not deviate from the distribution of random noise while preserving the quality of stego images and ensuring high extracted message accuracy.
We offer three types of message projection, which can be adjusted based on which problem to address.

\our{} does not require fine-tuning pre-trained models or training additional models such as extractors or decoders.
By using well-learned pre-trained models, \our{} generates high-quality stego images, FID score~\cite{heusel2017gans} of 2.77 on FFHQ 64$\times$64~\cite{karras2019style} images while hiding 1.0 bits per pixel (bpp) messages.
Additionally, using pre-trained diffusion models trained on AFHQv2 64$\times$64~\cite{choi2020starganv2}, \our{} achieves hiding 6.0 bpp messages with high extracted message accuracy. 
Furthermore, we show that \our{} can be easily applied to text-to-image models~\cite{ramesh2022hierarchical,saharia2022photorealistic}, such as Stable diffusion~\cite{rombach2022high}, by leveraging only secret messages and text prompts.

\section{Preliminaries}
\subsection{Generative Steganography}
In generative steganography, two players, the sender and the receiver, communicate secret messages $\mathbf{M}$ through generated stego image $\mathbf{X}_S$.
Unlike traditional steganography methods~\cite{morkel2005overview}, the sender in generative steganography uses a generator $G$ to generate $\mathbf{X}_S$ from $\mathbf{M}$ without cover image $\mathbf{X}_C$. 
The receiver extracts secret messages $\mathbf{M}'$ from $\mathbf{X}_S$ using an extractor $E$.
The process is defined as follows : 
\begin{align}   
G(\mathbf{M}) = \mathbf{X}_S, \quad E(\mathbf{X_S}) = \mathbf{M}'.
\end{align}

In generative steganography, both image quality and extracted message accuracy are significant.
Image quality is an indicator of how stego images are photorealistic like real images.
The stego images should be visually and statistically similar to the real images to deceive the steganalyzer.
Additionally, the generator needs to produce stego images in such a way that the receiver can extract the secret messages.
In \our{}, we utilize diffusion models as both a generator and an extractor of generative steganography.

\subsection{Diffusion Models}
\label{sec2_2}
Diffusion models~\cite{sohl2015deep,ho2020denoising} generate images through an iterative denoising process from Gaussian noise.
The sampling process can be viewed as solving the ODE process, with time $t$ going from $T$ to 0, using deterministic samplers~\cite{song2020denoising}.
This process is termed probability flow ODE~\cite{song2020score} and follows below equation:
\begin{align}
\label{eqn:pfode}
\mathrm{d}\mathbf{x} = [\mathbf{f}(\mathbf{x}, t) - \frac{1}{2} g(t)^2 \mathbf{s}_{\mathbf{\theta}}(\mathbf{x},t)]\mathrm{d}t,
\end{align}
where $\mathbf{f}$ is drift coefficient, $g$ is diffusion coefficient and $\mathbf{s}_{\mathbf{\theta}}$ is score function trained as neural network.
The score function estimates $\nabla_\mathbf{x}\text{log} p_t(\mathbf{x})$, where $p_t(\mathbf{x})$ is the probability distribution of $\mathbf{x}(t)$.

Through Equation~\eqref{eqn:pfode}, diffusion models generate the image $\mathbf{x}(0)$ from Gaussian noise $\mathbf{x}(T) = \sigma_T \mathbf{z}$, where $\sigma_T$ is constant of $T$ and $\mathbf{z}$ is standard Gaussian noise, $\mathbf{z} \sim \mathcal{N}(0,I)$.
We can establish a bijective function between $\mathbf{z}$ and image $\mathbf{x}(0)$ using ODE, namely, $\mathbf{x}(0) = f_\theta(\mathbf{z})$.
The function $f_\theta$ is invertible, we can express another equation $\mathbf{z} = f_\theta^{-1}(\mathbf{x}(0))$.

\begin{figure}[t]
    \centering
    \includegraphics[width = 0.98\textwidth]{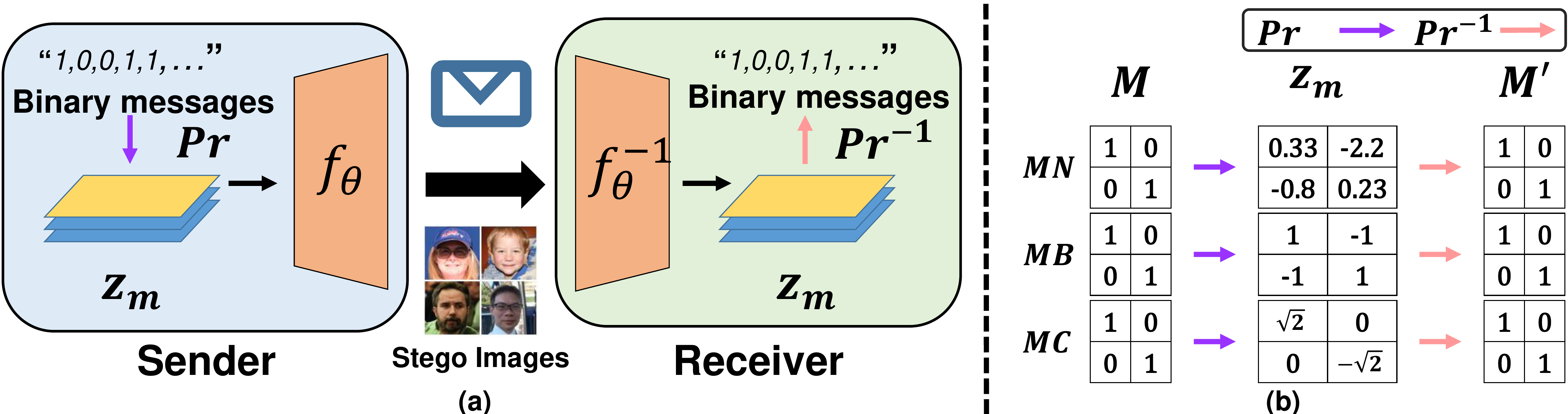}
    \caption{The overview of \our{}. (a) Generative steganography process of \our{}. (b) The example of three message projection processes when messages are `1001'.}
    \label{fig:ov}
    \vspace{-1em}
\end{figure}

\section{Method}
\subsection{\our{}}
\label{3.1}
\paragraph{Settings}

We propose \our{}, a new generative steganography framework leveraging pre-trained diffusion models.
\our{} can hide $n$ bpp binary messages, $\mathbf{M} \in \{ 0, 1 \}^{n\times W\times H}$ in images $\mathbf{X} \in \, \mathbb{Z}^{3\times W\times H}$, where $W$ and $H$ denote the width and height of images.

In \our{}, we consider two players, the sender and the receiver.
The sender projects the secret binary messages $\mathbf{M}$ into Gaussian noise and generates stego images $\mathbf{X}_S$ from the projected noise using a diffusion model.
Then, the receiver extracts the hidden noise using the same diffusion model and projects it onto the binary message.
We note that the sender and the receiver should share the same diffusion model and the projection process to use \our{}.
There are no restrictions on the shared diffusion models, so two players can use any pre-trained diffusion models.

\cite{wei2022generative2} demonstrated that saving images as float type using TIFF format resulted in higher performance in extracted message accuracy than saving as integer type using PNG or JPEG formats.
However, in our research, we mainly use integer type to save images because PNG and JPEG formats are more widely used than TIFF.
We will present our result in Section~\ref{sec4}, including the results of using the TIFF format.

\paragraph{Procedure}

Generative steganography requires two models: $G$, which generates images from messages, and $E$, which extracts messages from images. 
Previous works have trained two separate models for $G$ and $E$, similar to the $f_\theta$ and $f_\theta^{-1}$ described in Section~\ref{sec2_2}.
However, both models can be performed by a single diffusion model.
Thus, we can use diffusion models as generative steganography, provided that message projection projects $\mathbf{M}$ into the same domain as Gaussian noise $\mathbf{z}$. 
We use deterministic samplers such as DDIM sampler~\cite{song2020denoising} or Heun's sampler of EDM~\cite{karras2022elucidating} for the invertible function $f_\theta$.

In \our{}, we edit Gaussian noise $\mathbf{z}$ with \zm{} $\mathbf{z}_m$, where $\mathbf{z}_m$ is the noise hiding $\mathbf{M}$. 
The number of channels of $\mathbf{z}_m$ to hide messages depends on the message quantity.
When the length of messages is $n$ bpp, we use $n$ channels of $\mathbf{z}_m$.
Messages projection $Pr$ is function that maps $\mathbf{z}$ and $\mathbf{M}$ into $\mathbf{z}_m$, $\mathbf{z}_m = Pr(\mathbf{z}, \mathbf{M})$.
If $Pr$ is invertible, we can generate a stego image $\mathbf{X}_S$ and extract hidden messages $\mathbf{M}'$ by solving the ODE process, $\mathbf{x}_{s}(0) = f_\theta(Pr(\mathbf{z}, \mathbf{M}))$ and $\mathbf{M}' = Pr^{-1}(f^{-1}_\theta (\mathbf{x}_{s}(0)))$, as shown in Figure~\ref{fig:ov}(a).

\subsection{Challenging Problems of \our{}}
\label{problem}

This section introduces two challenging problems when using diffusion models for generative steganography.

\begin{figure}

  \centering
  \subfigure[]{\includegraphics[width = 0.245\textwidth]{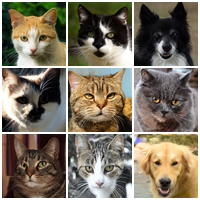}}
  \subfigure[]{\includegraphics[width = 0.245\textwidth]{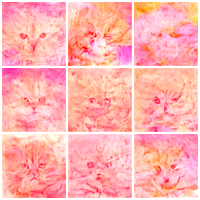}}
  \subfigure[]{\includegraphics[width = 0.245\textwidth]{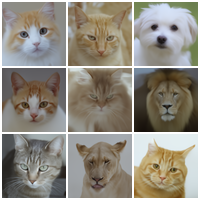}}
  \subfigure[]{\includegraphics[width = 0.245\textwidth]{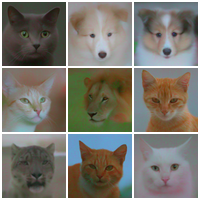}}
  \caption{Images generated by diffusion models. (a) Normally generated images. (b) Collapsed images hiding 1.0 bpp, mean of $\mathbf{z}_m$ differs from $\mathbf{z}$. (c) Collapsed images hiding 1.0 bpp, variance of $\mathbf{z}_m$ differs from $\mathbf{z}$. (d) Collapsed images hiding 1.0 bpp using se-S2IRT~\cite{zhou2022secret} algorithm (each value of $\mathbf{z}_m$ is not independent). Additional samples of image collapse are provided in Appendix~\ref{appendix_sample}}
  \label{fig:collapse}
  \vspace{-1em}
\end{figure}

\paragraph{Image collapse} 
In \our{}, we utilize the invertible property of the deterministic sampler of diffusion models.
Diffusion models have learned to map Gaussian noise to images.
If the projection $Pr$ is naively defined, the distribution of $\mathbf{z}_m$ can differ from that of Gaussian noise. 
In this case, image quality may be harmed or even collapse, as shown in Figure~\ref{fig:collapse}.
We refer to this issue as \textit{image collapse}.

To prevent the image collapse, we need to make the distribution of $\mathbf{z}_m$ similar to that of Gaussian noise.
In Figure~\ref{fig:collapse}, we show that $\mathbf{z}_m$ should satisfy following the three conditions:
\textbf{(1)} The \textbf{mean} of $\mathbf{z}_m$ is close to 0: if the mean is higher than 0, the output image becomes white, while if it is lower than 0, it becomes dark.
\textbf{(2)} The \textbf{variance }of $\mathbf{z}_m$ is similar to 1: when the variance is too high or too low, the output image may not be properly denoised or may be oversimplified.
\textbf{(3)} The values of $\mathbf{z}_m$ are \textbf{independent}: when the values are not independent, diffusion models do not work normally.

\begin{wrapfigure}{r}{0.38\textwidth}
    \vspace{-15pt}
    \includegraphics[width = 0.36\textwidth]{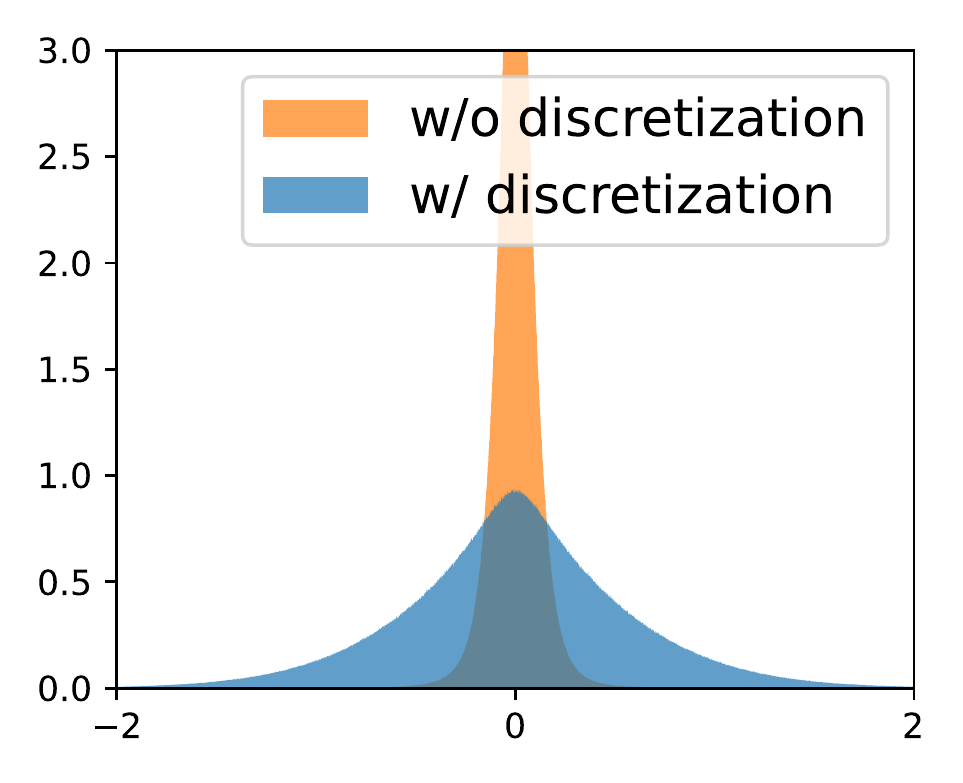}
    \caption{Density distribution of $\mathbf{z} - \mathbf{z}'$, where the error between the input noise $\mathbf{z}$ and the extracted noise $\mathbf{z}' = f_\theta^{-1}(f_\theta(\mathbf{z}))$. The orange line: the error without discretization. The blue line: the error with discretization into integers.}
    \vspace{-30pt}
    \label{fig:bel}
\end{wrapfigure}

\paragraph{Extraction error}
As the deterministic sampler solves the ODE process with a numerical integrator, errors accumulate in both the forward and backward processes of the ODE.
This can lead to a decrease in the extracted message accuracy, which we term as \textit{extraction error}.
Additionally, in the steganography process, the sender needs to save images in integer formats, such as PNG or JPEG, to send to the receiver.
Discretization during image saving amplifies the error. 
In Figure~\ref{fig:bel}, we show the accumulated error between input noise and extracted noise.
Small errors occur during the numerical integrator (the orange line), and the errors become larger due to discretization (the blue line).

\subsection{Message Projection}
\label{projection process}

In this section, we describe message projection, the key ingredient for solving both problems, the image collapse and the extraction error.
We suggest three options for message projection, which are designed depending on which problem to focus on.
Examples of our message projections are shown in Figure~\ref{fig:ov}(b).

\paragraph{Message to Noise (MN)}

We propose MN projection $Pr_{N}$ to solve the image collapse.
The projection $Pr_{N}$ maps the messages to $\mathbf{z} \sim \mathcal{N}(0,I)$, so that the distribution of \zm{} $\mathbf{z}_m$ is equivalent to that of Gaussian noise.

To implement the MN projection, we first sample standard Gaussian noise $\mathbf{z}$. 
Then, we project $\mathbf{z}$ into $\mathbf{z}_m$ using the following rule: change the sign of $\mathbf{z}$ to a positive number where $\mathbf{M}$ is 1, and to a negative number where $\mathbf{M}$ is 0.
The receiver can extract the messages by applying the inverse projection $Pr_{N}^{-1}$, which checks whether the value of extracted $\mathbf{z}_m$ is greater than 0 or not.

Since the probability distribution of $\mathbf{z}_m$ is the same as that of the Gaussian noise, generated images $f_\theta(Pr_{N}(\mathbf{z},\mathbf{M}))$ are challenging to distinguish from normally generated images.
However, using MN projection is vulnerable to the extraction error because some values of $\mathbf{z}_m$ are situated near the decision boundary of $Pr_{N}^{-1}$, which is 0.
Therefore, we suggest other projection options to improve the extracted message accuracy.

\paragraph{Message to Binary (MB)}

The MB projection is designed to solve the aforementioned problem, the extraction error.
In the MB projection, the receiver identifies the messages by inverse projection $Pr_{B}^{-1}$, the same as that of the MN projection. $Pr_{N}^{-1}$.
The MB projection $Pr_{B}$ maps the values of $\mathbf{z}_m$ as far as possible from 0, which corresponds to the decision boundary of $Pr_{B}^{-1}$. 

To maximize the minimum distance from the boundary, $Pr_{B}$ equals all the values that denote the same messages.
To ensure that the variance of $\mathbf{z}_m$ becomes 1, $Pr_{B}$ set the value of $\mathbf{z}_m$ to 1 where $\mathbf{M}$ is 1 and to -1 where $\mathbf{M}$ is 0.

\paragraph{Message to Centered Binary (MC)}

While the MN projection resolves the image collapse, it performs worse in terms of extracted message accuracy than the MB projection.
The MB projection well addresses the extraction error. 
However, the distribution of $\mathbf{z}_m$ deviates from that of Gaussian noise, which causes a slight degradation in image quality. 
Therefore, we suggest a compromise between the two projections, called the MC projection.

Similar to the MB projection, MC projection $Pr_{C}$ projects the values which denote the same messages into coherent values.
Since the mode of Gaussian noise is 0, we set the value of $\mathbf{z}_m$ to 0, where $\mathbf{M}$ is 0.
When $\mathbf{M}$ is 1, we randomly map the value of $\mathbf{z}_m$ to either $\sqrt{2}$ or $-\sqrt{2}$, so that the mean and variance of $\mathbf{z}_m$ are equal to those of Gaussian noise.
Inverse projection $Pr_{C}^{-1}$ checks the values of $\mathbf{z}_m$ are close to $\sqrt{2}$, $-\sqrt{2}$, or 0.
 
The MC projection performs higher extracted message accuracy than the MN projection and better sample quality than the MB projection, which will be demonstrated in Section~\ref{sec4}.

\subsection{Trick of Hiding Large Messages}

Generally, input noise for diffusion models consists of 3 channels.
When applying the projections referred to in Section~\ref{projection process}, the maximum capacity of secret messages is 3.0 bpp. 
In order to conceal more messages than 3.0 bpp, we should hide more than 1.0 bpp messages in a single channel.

We hide multiple bits following MB, which we call Multi-bits projection.
Two bits messages consist of four cases: 00, 01, 10, and 11.
We set four values and keep them as far away from each other as possible while maintaining the mean and variance of $\mathbf{z}_m$.
For 2 bits, we define the value as $-3/\sqrt{5}, -1/\sqrt{5}, 3/\sqrt{5}$, or $1\sqrt{5}$, where $\mathbf{M}$ is 00, 01, 10, or 11.
We can hide 6.0 bpp messages by hiding 2 bits in each channel and more messages by applying this projection.

\section{Experiments}
\label{sec4}

\paragraph{Datasets and pre-trained models}
We consider three commonly used image datasets for generative models: 
CIFAR-10~\cite{krizhevsky2009learning}, FFHQ 64$\times$64~\cite{karras2019style} and AFHQv2 64$\times$64~\cite{choi2020starganv2}.
Several previous works~\cite{song2020score,karras2022elucidating,xu2022poisson,xu2023pfgm++} provide pre-trained models trained on these datasets.
In our experiments, we use pre-trained models and deterministic Heun's sampler of EDM~\cite{karras2022elucidating}.
For the FFHQ 64$\times$64 and AFHQv2 64$\times$64 datasets, we processed $f_\theta$ and $f_\theta^{-1}$ with 40 inference steps, while for the CIFAR-10 dataset, we used 18 inference steps.
We conduct our experiments using 4 Nvidia Titan Xp GPUs.

\paragraph{Metrics}
We evaluate extracted message accuracy, anti-detection ability, and image quality of our methods.
The accuracy (Acc) measures the accuracy of the extracted message, which may be distorted through the steganography process $f_\theta$ and $f_\theta^{-1}$.
We calculate Acc as follows: Acc $ = 1 - \frac{\mathbf{M} \oplus \mathbf{M}'}{\mathrm{len}(\mathbf{M})}$, where $\mathbf{M}$ is original binary messages, $\mathbf{M}'$ is extracted binary messages through the steganography process, $\oplus$ is XOR operator, and $\mathrm{len}(\mathbf{M})$ is length of the messages.
The detection error (Pe) is an indicator of the performance of classifier models.
Pe is defined as follows: Pe $ = \mathrm{min}_{P_{FA}} \frac{1}{2}(P_{FA} + P_{MD})$, where $P_{FA}$ and $P_{MD}$ are the rates of false-alarm and miss-detection errors. A Pe value of 0.5 means that the classifier can not distinguish two classes completely.
We use Xu-Net~\cite{xu2016structural} models to evaluate Pe of steganalyzer models and assess the anti-detection ability of our method.
Frechet inception distance (FID) score~\cite{heusel2017gans} is an image quality assessment indicator, where a lower FID score indicates better image quality.
Bit per pixel (bpp) is a unit for the message quantity hiding in images. 
We calculate bpp as follows: $\frac{\mathrm{len}(\mathbf{M})}{W \times H}$,where $W$ and $H$ are width and height of image.

In our experiments, We sampled 6,000 stego images from each model to calculate the accuracy.
We divide the stego images into 5,000 training sets and 1,000 test sets for training and evaluating steganalyzer models. 
We train Xu-Net on 5,000 stego images and 5,000 real images and test on 1,000 stego images and 1,000 real images for each steganography model, following in \cite{zhou2022secret}.
We sample 50,000 stego images that hide random messages to calculate FID scores.

\paragraph{Steganalyzer settings}

As generative steganography models do not have cover images, third-party players cannot train their steganalyzer models.
Therefore, we utilize the real images as a substitute of cover images, assuming that third-party players would adopt the strict strategy in their steganalysis.

\paragraph{Baseline}

We select two baseline generative steganography models which can hide messages above 1.0 bpp, Generative Steganography Network (GSN)~\cite{wei2022generative} and Secret to Image Reversible Transformation (S2IRT)~\cite{zhou2022secret}.

GSN is a GAN-based~\cite{goodfellow2020generative} method that consists of four models: generator, discriminator, steganalyzer, and extractor.
In our experiments, we train each GSN model from scratch with different payload settings.
S2IRT is a generative steganography method that applies Glow~\cite{kingma2018glow} models.
We train Glow and use Separate Encoder based S2IRT (SE-S2IRT) scheme.
The SE-S2IRT scheme splits random values into $K$ clusters and assigns each message to the corresponding cluster based on the order of values.
Increasing $K$ leads to a higher message capacity but lower accuracy.
In our experiments, we choose a low value of $K$ to optimize extracted message accuracy.

\paragraph{Discretization}

In our experiments, we discretize stego images into integer values and save them in PNG format.
\cite{wei2022generative2} proposed that using the TIFF format shows good performance in terms of extracted message accuracy because TIFF format saves an image in continuous values.
Following~\cite{wei2022generative2}, we also conduct experiments using TIFF format and present the results in Section~\ref{tiff}.

\begin{table}[t]
    \caption{Comparison of extracted message accuracy (Acc, \%), anti-detection ability (Pe), and image quality (FID) with baseline methods. $^\dagger$: our re-implementation.}
    \label{baselines}
    \centering
    {
    \begin{tabular}{cccccccccc}
    \toprule
       & \multicolumn{3}{c}{1.0 bpp}  & \multicolumn{3}{c}{2.0 bpp} & \multicolumn{3}{c}{3.0 bpp}\\
    \midrule
    Method & \multicolumn{1}{c}{Acc $\uparrow$} & \multicolumn{1}{c}{Pe $\uparrow$} & \multicolumn{1}{c}{FID $\downarrow$} & \multicolumn{1}{c}{Acc $\uparrow$} & \multicolumn{1}{c}{Pe $\uparrow$} & \multicolumn{1}{c}{FID $\downarrow$} & \multicolumn{1}{c}{Acc $\uparrow$} & \multicolumn{1}{c}{Pe $\uparrow$} & \multicolumn{1}{c}{FID $\downarrow$} \\\midrule
    GSN$^\dagger$      & 97.15& 0.183& 13.4& 79.62& 0.022& 24.8& 72.74& 0.049& 30.4 \\ 
    S2IRT$^\dagger$ & \textbf{99.94}& 0.003& 72.6& 97.79& 0.002& 78.2& 97.13& 0.003& 67.8 \\ 
    \our{}       & 98.12& \textbf{0.427}& \textbf{2.77}& \textbf{98.19}& \textbf{0.361}& \textbf{3.30}& \textbf{98.76}& \textbf{0.310}& \textbf{4.30} \\ 
 \bottomrule                              
    \end{tabular}
    }
    \vspace{-1em}
\end{table}

\begin{figure}[t]
    \centering
    \subfigure[]{\includegraphics[width = 0.32\textwidth]{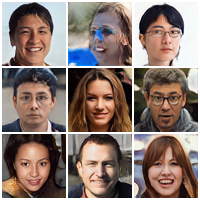}}
    \subfigure[]{\includegraphics[width = 0.32\textwidth]{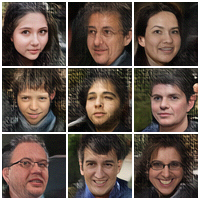}}
    \subfigure[]{\includegraphics[width = 0.32\textwidth]{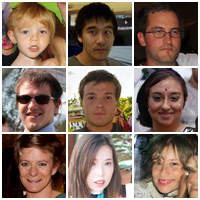}}
    \caption{FFHQ 64$\times$64 stego images hiding 1.0 bpp messages. (a) GSN, (b) S2IRT, (c) \our{}. Stego images generated by \our{} show higher quality than the other baseline methods.}
    \label{fig:ffhq_baseline}
    \vspace{-1em}
\end{figure}

\subsection{Comparison with Baseline Models.}

We compare \our{} with two high-capacity generative steganography models, GSN and S2IRT, which are trained on the FFHQ 64$\times$64 dataset.
 The comparison is performed using the MB projection in three payload settings: 1.0 bpp, 2.0 bpp, and 3.0 bpp. 

The results are shown in Table~\ref{baselines}.
The results presented in Figure~\ref{fig:ffhq_baseline} and Table~\ref{baselines} show that \our{} outperforms the other baseline models in terms of anti-detection ability and image quality.
S2IRT shows higher accuracy than the other two models when the message capacity of stego images is 1.0 bpp. 
However, when the message capacity is higher than 1.0 bpp, \our{} showed higher accuracy than S2IRT. 
Although S2IRT achieves high accuracy (99.94\% at 1.0 bpp) when the hyper-parameter of S2IRT $K$ is 2, its message capacity is limited to 1.5 bpp.
To hide 2.0 bpp messages, $K$ should be at least 3, which decreases the extracted message accuracy.
GSN shows competitive accuracy in the payload setting of 1.0 bpp, but it decreases rapidly as the payload increases.

\subsection{Ablation Study}

We compare the performance of our message projection options in our experiments using the FFHQ and AFHQv2 datasets.
The results of our comparison are presented in Table~\ref{table_1-3bppn123}. 
Using the MB projection outperforms the other two projections in Acc.
When hiding small messages with payloads of 1.0 bpp, the Pe and FID scores of each projection are similar. 
However, hiding large messages, such as with payloads of 3.0 bpp, the anti-detection ability and image quality of using the MB projection decreases rapidly compared to using the MN projection.
In the FFHQ dataset, the MC projection shows compromised results of Acc, Pe, and FID between the MN projection and the MC projection as the payload of messages increases.

\begin{table}[t]
    \caption{Ablation study results of three projection options on FFHQ 64$\times$64 and AFHQv2 64$\times$64 datasets. The original EDM models have FID scores of 2.39 on FFHQ and 2.17 on AFHQv2.}
    \label{table_1-3bppn123}
    \centering
    {\footnotesize
    \begin{tabular}{cc|ccc|ccc|ccc}
    \toprule
    \multirow{2}{*}{Datasets} &\multirow{2}{*}{Projections}    & \multicolumn{3}{c}{1.0 bpp}  & \multicolumn{3}{|c}{2.0 bpp} & \multicolumn{3}{|c}{3.0 bpp}\\
    && \multicolumn{1}{c}{Acc $\uparrow$} & \multicolumn{1}{c}{Pe $\uparrow$} & \multicolumn{1}{c}{FID $\downarrow$} & \multicolumn{1}{|c}{Acc $\uparrow$} & \multicolumn{1}{c}{Pe $\uparrow$} & \multicolumn{1}{c}{FID $\downarrow$} & \multicolumn{1}{|c}{Acc $\uparrow$} & \multicolumn{1}{c}{Pe $\uparrow$} & \multicolumn{1}{c}{FID $\downarrow$} \\\midrule
    \multirow{3}{*}{FFHQ} & MN       & 88.00& 0.422& \textbf{2.41}& 86.75& \textbf{0.433}& \textbf{2.42}& 87.06& \textbf{0.427}& \textbf{2.45} \\
    &MB       & \textbf{98.12}& 0.427& 2.77& \textbf{98.19}& 0.361& 3.30& \textbf{98.76}& 0.310& 4.30 \\
    &MC       & 93.17& \textbf{0.445}& 2.58& 91.97& 0.414& 2.75& 93.09& 0.409& 3.11 \\ \midrule
    \multirow{3}{*}{AFHQv2}&MN     & 87.32& 0.399& \textbf{2.14}& 85.68& 0.390& \textbf{2.20}& 86.64& 0.403& \textbf{2.13} \\ 
    &MB     & \textbf{98.03}& 0.396 & 2.21& \textbf{98.57}& 0.388 & 2.35& \textbf{99.19}& 0.376& 2.46 \\ 
    &MC     & 92.65& \textbf{0.407}& 2.22& 91.00& \textbf{0.404}& 2.26& 93.40& \textbf{0.405}& 2.26 \\
    \bottomrule                              
    \end{tabular}
    }
    \vspace{-1em}
\end{table}

\subsection{Performance of Hiding High Bpp Messages}
\label{high}

\begin{figure}[t]
    \centering
    \subfigure[1.0 bpp]{\includegraphics[width = 0.16\textwidth]{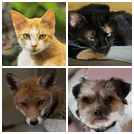}}
    \subfigure[2.0 bpp]{\includegraphics[width = 0.16\textwidth]{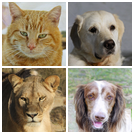}}
    \subfigure[3.0 bpp]{\includegraphics[width = 0.16\textwidth]{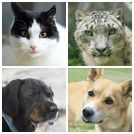}}
    \subfigure[4.0 bpp]{\includegraphics[width = 0.16\textwidth]{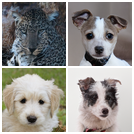}}
    \subfigure[5.0 bpp]{\includegraphics[width = 0.16\textwidth]{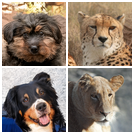}}
    \subfigure[6.0 bpp]{\includegraphics[width = 0.16\textwidth]{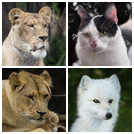}}
    \caption{AFHQv2 64$\times$64 images which are hiding messages with pre-trained EDM models. \our{} can hide 6.0 bpp messages without the image collapse.}
    \label{fig:afhq_6bpp}
    \vspace{-1em}
\end{figure}

We evaluate \our{} on various payload settings, ranging from 1.0 bpp to 6.0 bpp for each dataset.
In payloads from 1.0 to 3.0 bpp, we use the MB projection. 
In payloads from 4.0 and 5.0 bpp, we use both the MB projection and the Multi-bits projection.
In 6.0 bpp payloads, we only use the Multi-bits projection.

Figure~\ref{fig:afhq_6bpp} shows the stego images generated by AFHQv2 models.
As shown in Table~\ref{table_1-6bpp}, using the MB projection alone (from 1.0 to 3.0 bpp) results in higher extracted message accuracy compared to using the Multi-bits projection.
However, using the Multi-bits projection (from 4.0 to 6.0 bpp) provides better anti-detection ability and image quality. 
This is because the distribution of $\mathbf{z}_m$ projected by the Multi-bits projection is more similar to that of Gaussian noise than that of the MB projection.
When the model is trained on the CIFAR-10 dataset, the extracted message accuracy is lower than those trained on other datasets. 
This is due to the susceptibility of generated CIFAR-10 images to discretization, which will be presented in Section~\ref{tiff}.

As the number of bits to hide in channels increases, the extracted message accuracy decreases due to the same reason as the MN projection.
We hide 9.0 bpp messages using EDM models trained on the AFHQv2 dataset and the Multi-bits projection.
The extracted message accuracy from the stego images hiding 9.0 bpp messages is 83.18\%.

\begin{table}[t]
    \caption{Results of different message payloads, 1.0 to 6.0 bpp. The original EDM models have FID scores of 2.39 on FFHQ, 2.17 on AFHQv2, and FID scores of 1.97 on CIFAR-10. }
    \label{table_1-6bpp}
    \centering
    {
    \begin{tabular}{cccccccc}
    \toprule
    
     Datasets               & metric & 1.0 & 2.0 & 3.0 & 4.0 & 5.0 & 6.0  \\ 
                            \midrule
    \multirow{3}{*}{AFHQv2 64$\times$64} 
                            & Acc $\uparrow$ & 98.03 & 98.57 & 99.19 & 96.39 & 93.15 & 91.93 \\
                            & Pe $\uparrow$ & 0.396 & 0.388 & 0.376 & 0.364 & 0.390 & 0.394 \\
                            & FID $\downarrow$ & 2.21 & 2.35 & 2.46 & 2.42 & 2.35 & 2.34 \\ 
                            \midrule
    \multirow{3}{*}{FFHQ 64$\times$64}
                            & Acc $\uparrow$ & 98.12 & 98.19 & 98.76 & 95.57 & 92.38 & 91.12 \\
                            & Pe $\uparrow$ & 0.427 & 0.361 & 0.310 & 0.334 & 0.345 & 0.385
 \\
                            & FID $\downarrow$ & 2.77 & 3.30 & 4.30 & 3.90 & 3.67 & 3.37 \\ 
                            \midrule
    \multirow{3}{*}{CIFAR-10} & Acc $\uparrow$& 95.38 & 95.07 & 95.19 & 89.93 & 86.43 & 84.83 \\
                            & Pe $\uparrow$ & 0.434 & 0.460 & 0.434 & 0.417 & 0.446 & 0.441 \\
                            & FID $\downarrow$ & 2.09 & 2.30 & 2.66 & 2.48 & 2.37 & 2.31 \\ 
    \bottomrule
    \end{tabular}
    }
    \vspace{-1em}
\end{table}

\subsection{Applying on Pre-Trained Text-to-Image Models}

\begin{figure}[t]
    \centering
    \subfigure[]{\includegraphics[width = 0.24\textwidth]{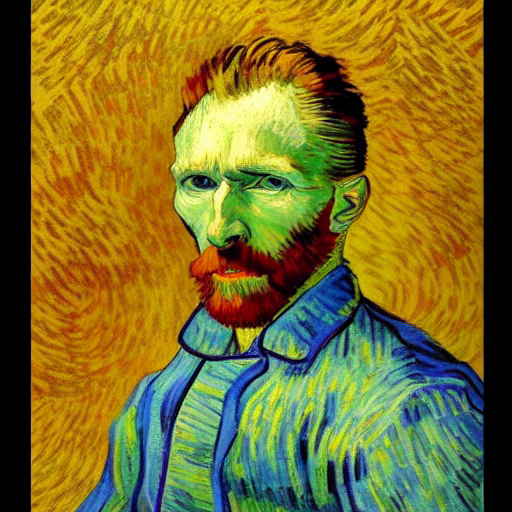}}
    \subfigure[]{\includegraphics[width = 0.24\textwidth]{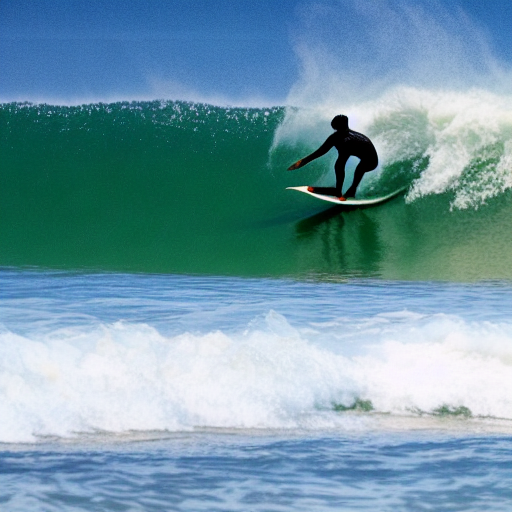}}
    \subfigure[]{\includegraphics[width = 0.24\textwidth]{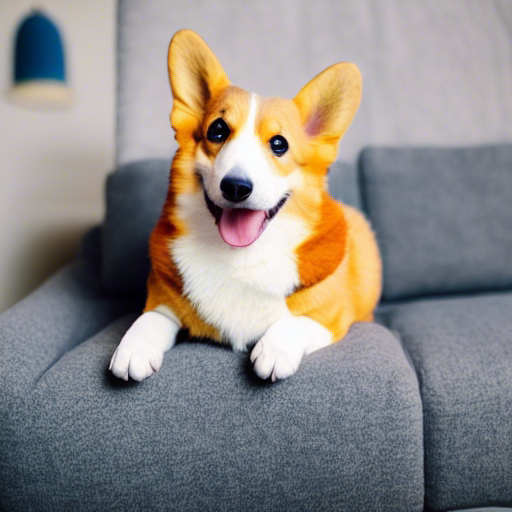}}
    \subfigure[]{\includegraphics[width = 0.24\textwidth]{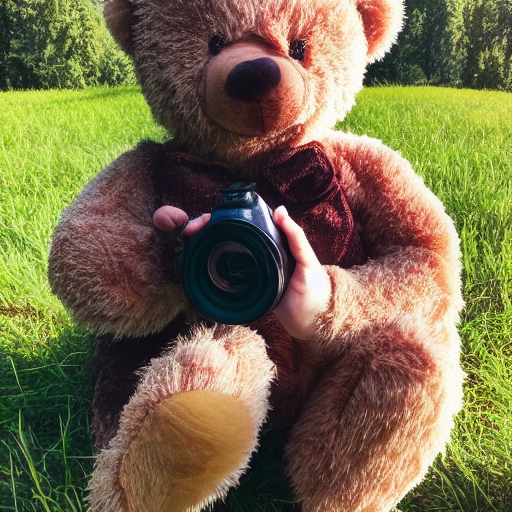}}
    \caption{Sample stego images generated by Stable diffusion. Stable diffusion can generate high-quality images while hiding secret messages. (a) `A painting of Gogh' (Acc: 92.09\%). (b) `A photograph of a surfer' (Acc: 98.25\%).  (c) `A photograph of a corgi sitting on a couch' (Acc: 97.04\%). (d) `A photograph of a teddy bear taking a photo' (Acc: 98.04\%).}
    \label{fig:stablediffusion}
    \vspace{-1em}
\end{figure}

Our method can be extended to large-scale text-to-image models, which generate images with text guidance.
In this case, the sender and the receiver should share two additional pieces of information: input text prompt and guidance scale. 
We use Stable diffusion~\cite{rombach2022high}, an open-source model trained in the latent space of VAE~\cite{kingma2013auto}.
During inference, the diffusion model generates 4$\times$64$\times$64 latent features, which are then decoded to 512$\times$512 size images with VAE.
Thus, \our{} can conceal 0.0625 bpp messages when using Stable diffusion.
Figure~\ref{fig:stablediffusion} shows the samples generated by Stable diffusion using the MB projection.

\subsection{Message Accuracy According to Image Format}
\label{tiff}

We confirm the performance of the extracted message accuracy for different image formats by generating 1,000 images in each setting and calculating the accuracy (Acc) for each format. The projection methods used in these experiments are the same as those described in Section~\ref{high}.

The results, presented in Table~\ref{tifftable}, indicate that using the TIFF format results in a higher extracted message accuracy, as previously proposed in \cite{wei2022generative2}.

\begin{table}[t]

\centering
\caption{The extracted message accuracy from 1000 images with quantization to each image format.
Saving the images in TIFF format shows higher accuracy compared to the PNG format.
}
{
\begin{tabular}{cccccc}
\toprule
\multirow{2}{*}{Settings} & \multicolumn{2}{c}{\our{}} & S2IRT & GSN \\ 
      & CIFAR-10 3 bpp & FFHQ 6 bpp    & FFHQ 3 bpp        & FFHQ 1 bpp \\ \midrule 
Without quantize    & 100.00    & 99.77      & 100.00        & 99.60 \\ 
+ Quantize to TIFF   & 99.99     & 99.69      & 100.00      & 99.60 \\ 
+ Quantize to PNG     & 94.78    &  91.24   & 97.13        & 97.05 \\ \bottomrule
\end{tabular}
}
\vspace{-1em}
\label{tifftable}
\end{table}

\section{Related Work}

\subsection{Generative Steganography}

Generative steganography is a method where generative models synthesize images from secret messages without using any cover images.
Generative steganography offers several advantages over traditional steganography methods that use cover images.
One of the main benefits is that it can avoid detection by steganalysis methods because it does not modify images.
Further, steganalysis methods that are trained on such images cannot detect the presence of hidden data.

Early studies of generative steganography hide messages in simple images, such as texture or fingerprint images.
\cite{wu2014steganography} and~\cite{xu2015hidden} proposed approaches to hide secret messages in texture messages.
\cite{li2018toward} proposed the method to use fingerprint images.
These approaches generate low-quality and unnatural images, which are prone to be detected by third-party players.

Steganography approaches using generative models have been proposed to make high-quality and natural stego images. 
Especially, GAN models~\cite{goodfellow2020generative} have been used for generative steganography.
\cite{liu2017coverless} and~\cite{zhang2020generative} hide messages in label embedding of conditional GANs~\cite{mirza2014conditional, odena2017conditional}, \cite{hu2018novel, yu2021improved, wang2018sstegan, wei2022generative} train new extractor models.
\cite{zhou2022secret,wei2022generative2} proposed an approach to use invertible Flow models~\cite{kingma2018glow} to enable high capacity of hidden messages.

\subsection{Diffusion Models}

Diffusion models~\cite{sohl2015deep, ho2020denoising,song2020score} generate images through the stochastic iterative process of denoising from Gaussian noise. 
While this process incurs a high computational cost, it enables the generation of high-quality images.
Several studies~\cite{song2020denoising,song2020score} have proposed deterministic sampling methods for diffusion models. 
These methods aim to remove the stochastic property of diffusion models while sampling images using an invertible process.
\cite{song2020denoising} proposed an implicit sampling by changing the diffusion process to a non-Markov process.
\cite{song2020score} proposed probability flow ODE, which considers sampling processes as ODEs.
\cite{liu2022pseudo,zhang2022fast,lu2022dpm, karras2022elucidating} solve probability flow ODE efficiently using high order numerical integrator.
In our approach, we utilize Heun's sampler from the EDM~\cite{karras2022elucidating} to take advantage of the invertible property of the deterministic sampler.

\section{Conclusion}
We propose \our{}, a novel approach to generative steganography using deterministic samplers of diffusion models.
We investigate the factors that affect the quality and extracted message accuracy when diffusion models generate stego images.
We suggest three options for message projection, which have the trade-off of image quality, anti-detection, and extracted message accuracy.
Our approach can hide large messages (more than 1.0 bpp, even 6.0 bpp with an accuracy of 90\%) while maintaining the image quality of pre-trained diffusion models.

Our limitations include a trade-off between image quality, anti-detection ability, and extracted message accuracy. 
Our observations will enable future research to enhance these capabilities. 
Furthermore, since our approach uses pre-trained diffusion models, it can be extended to other domains such as video~\cite{ho2022video}, sound~\cite{kong2020diffwave}, and text~\cite{li2022diffusion}.

We are aware of the potential for \our{} to be exploited in information security threats.
Further details regarding the social impact of \our{} are provided in the Appendix~\ref{appendix_impact}.

{
\footnotesize
\bibliographystyle{plain}
\bibliography{egbib}
}

\newpage
\appendix

\section{Sample of Image Collapse}
\label{appendix_sample}
\begin{figure}[]
    \centering
    \includegraphics[width = 0.8\textwidth]{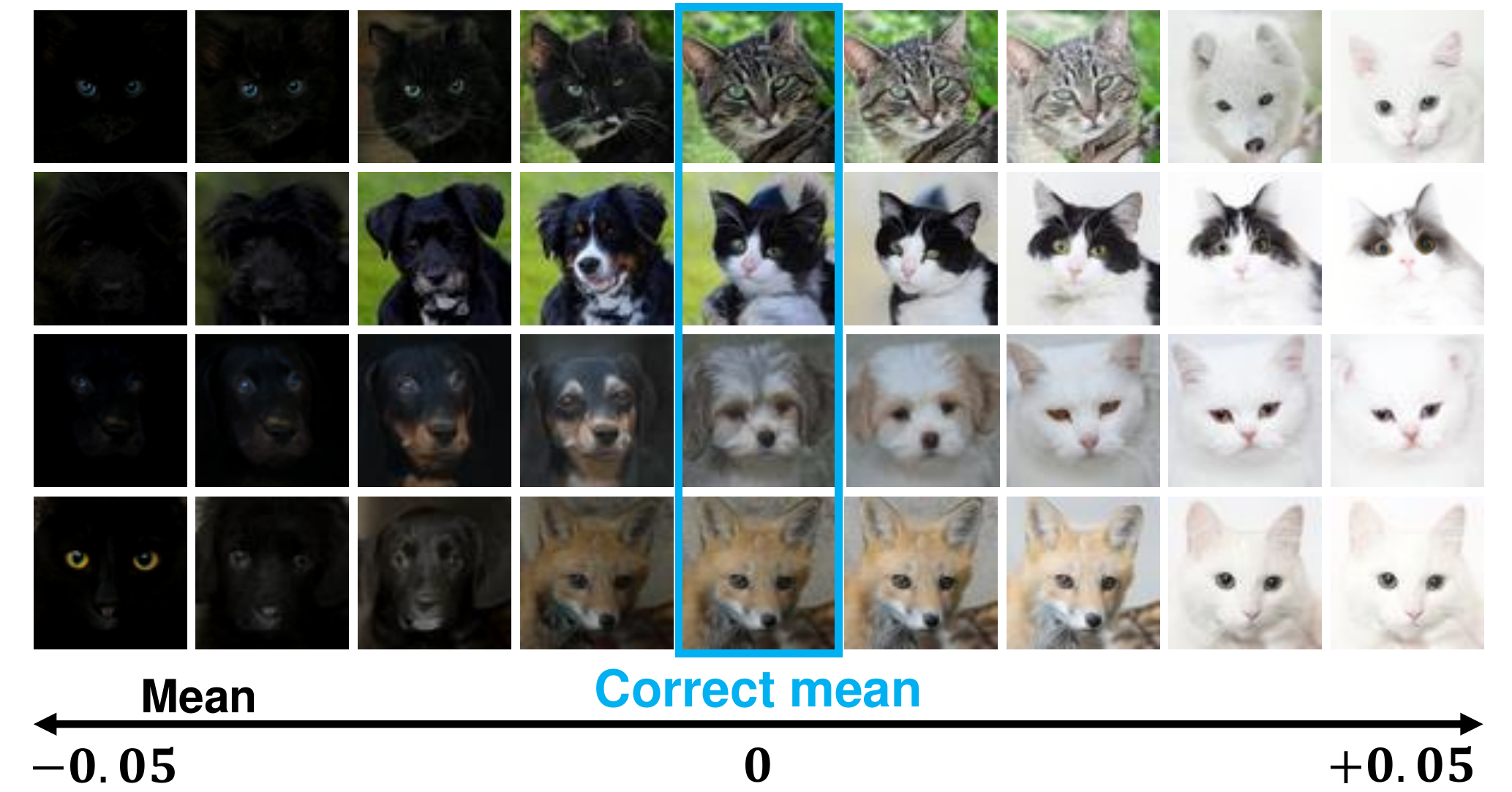}
    \caption{The influence of the mean value on stego images.}
    \label{app_co_mean}
\end{figure}

\begin{figure}[]
    \centering
    \includegraphics[width = 0.8\textwidth]{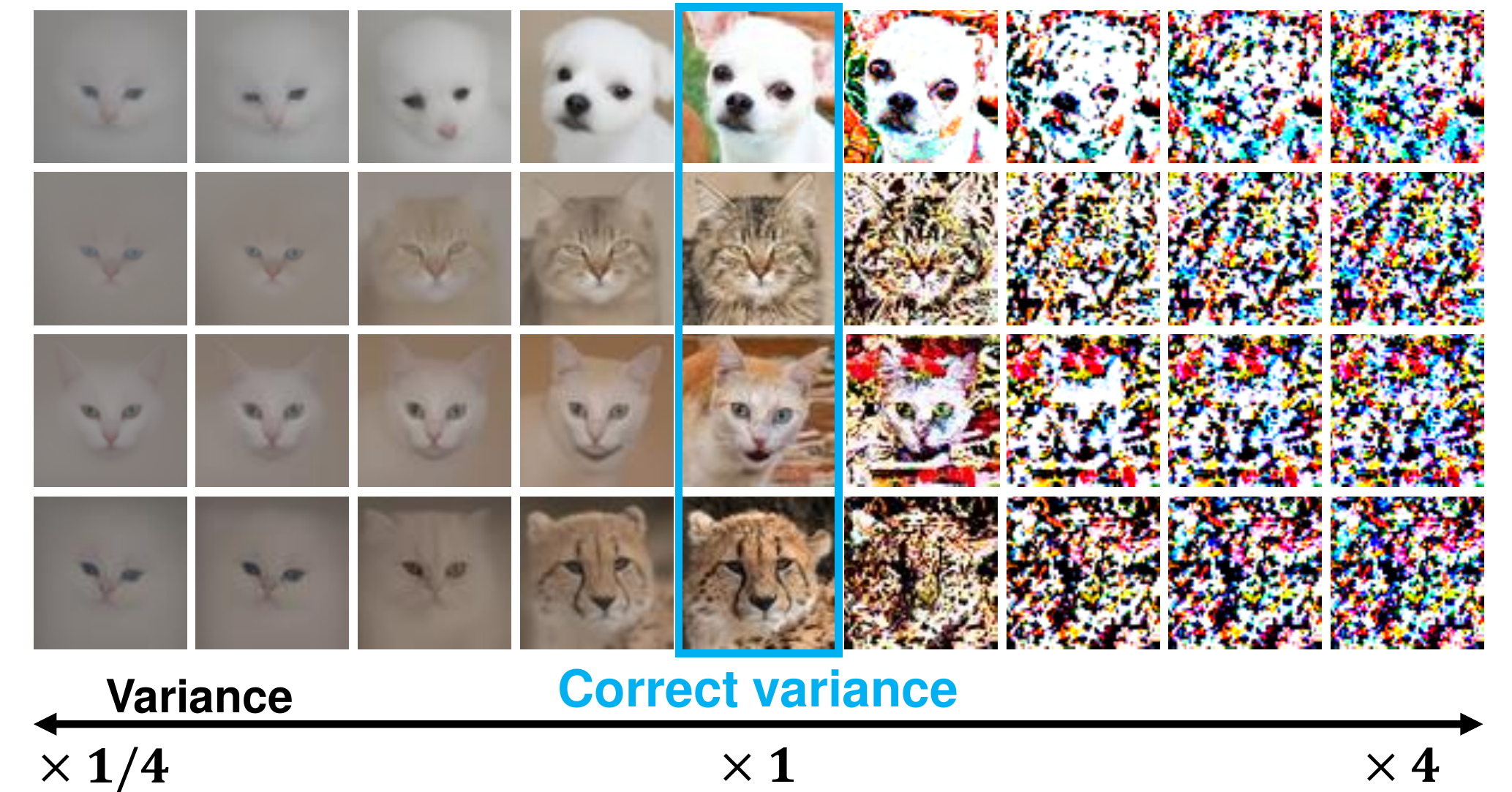}
    \caption{The influence of the variance value on stego images.}
    \label{app_co_var}
\end{figure}

In this section, we present examples where image collapse occurs. 
Figure~\ref{app_co_mean} illustrates stego images based on the mean of $\mathbf{z}_m$. 
When the mean is 0, similar to that of Gaussian noise, the diffusion model successfully generates high-quality images.
However, when the mean is higher or lower than 0, the images become brighter or darker, respectively. 
Figure~\ref{app_co_var} showcases stego images based on the variance of $\mathbf{z}_m$. 
When the variance is 1, similar to that of Gaussian noise, the diffusion model successfully generates high-quality images, too.
However, when the variance is high, the models fail to generate meaningful images, and when the variance is low, the images are oversimplified. 
Figure~\ref{app_co_ind} demonstrates three cases where the values of $\mathbf{z}_m$ are not independent.

\section{Trick for Higher Accuracy}

We introduce an additional projection called multi-channels projection, which enables the hiding of small messages with higher accuracy compared to other projections.

\paragraph{Settings}
In this approach, two players, the sender and the receiver, are required to share additional information compared to existing projections. Unlike the projections mentioned in the main paper, they need to share a binary codebook, $\mathbf{C} \in \{ 0, 1 \}^{3\times W\times H}$.

\begin{figure}[]
    \centering
    \subfigure[]{\includegraphics[width = 0.26\textwidth]{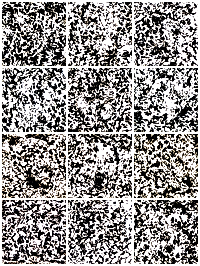}}
    \subfigure[]{\includegraphics[width = 0.26\textwidth]{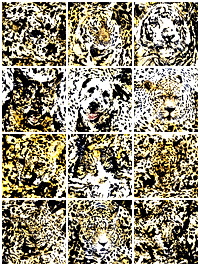}}
    \subfigure[]{\includegraphics[width = 0.26\textwidth]{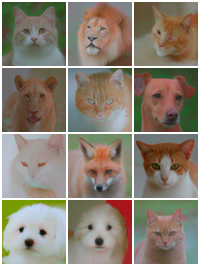}}
    \caption{Three examples of stego images where the values of $\mathbf{z}_m$ are not independent. (a) The values within the same pixel share the same sign. (b) The inputs of first and second channel are the same. (c) The algorithm of se-S2IRT with hyper-parameter $K$ = 2. }
    \label{app_co_ind}
\end{figure}

\paragraph{Multi-channels projection}
To improve accuracy, the sender hides multiple copies of the same message within a single image following the MB projection, and sends it. 
However, in this scenario, the values of $\mathbf{z}_m$ are not independent, leading to image collapse as shown in Figure~\ref{app_co_ind}(a). 
To address this issue, we utilize a codebook that is independent of the messages, ensuring the independence of $\mathbf{z}_m$ values. 
Before generating stego images, the sender modifies the sign of $\mathbf{z}_m$ where C is 0. 
Similarly, the receiver also modifies the sign of the extracted $\mathbf{z}_m$ where C is 0.
By checking the messages multiple times, the receiver achieves higher accuracy.

\paragraph{Experiments}
We use the same experimental settings as described in Section~\ref{sec4}. 
We consider two cases for the multi-channel projection. 
In the first case, both players use the same codebook, while in the second case, the players change their codebook for each stego image generated.
We hide messages with a capacity of 1.0 bpp, and the messages are replicated to three channels.

The results of the multi-channel projection are presented in Table~\ref{multichannel}. 
We observe that using the multi-channel projection yields higher accuracy compared to the MB projection when hiding 1.0 bpp messages.
However, when the codebook is not changed, the anti-detection ability of the multi-channel projection decreases, likely due to the similarity among images generated with the same codebook.

\section{Message Accuracy According to Image Format}
In this section, we present the accuracy results based on quantization, which serves as an extension of the experiments discussed in Section~\ref{tiff}. 
We evaluate the accuracy for different datasets and all of our projection techniques, as shown in Table~\ref{tiff_table}.

\begin{table}[b]
\centering
\caption{Results of the MB projection and multi-channel projection hiding 1.0 bpp messages. $^*$: changing the codebook  for each sampling.}
\begin{tabular}{ccccc}
\toprule
Datasets             & Projection & Acc $\uparrow$ & Pe $\uparrow$ & Fid $\downarrow$ \\
\midrule
\multirow{3}{*}{FFHQ} & MB (1.0 bpp) &98.12& 0.427& 2.77 \\
                     & Multi-channel & 99.88 & 0.274 & 2.42 \\
                     & Multi-channel$^*$ & 99.81 & 0.307 & 2.46 \\
                     \midrule
\multirow{3}{*}{AFHQ} & MB (1.0 bpp) & 98.03 & 0.396 & 2.21 \\
                     & Multi-channel & 99.76 & 0.321 & 4.43 \\
                     & Multi-channel$^*$ & 99.71 & 0.395 & 4.18 \\ \bottomrule
\end{tabular}
\label{multichannel}
\end{table}

\begin{table}[]
\centering
\caption{The extracted message accuracy from 1000 images. Settings of the experiments remain the same in Section~\ref{tiff}}
\begin{tabular}{ccccc}
\toprule
Datasets              & Projection & w/o quantization & To TIFF & To PNG \\
\midrule
\multirow{13}{*}{AFHQ} & MN (1.0 bpp)          & 97.71 & 97.62 & 87.35 \\
                      & MN (2.0 bpp)          & 97.39 & 97.28 & 85.80 \\
                      & MN (3.0 bpp)          & 97.55 & 97.45 & 86.52 \\
                      & MB (1.0 bpp)          & 100.00 & 100.00 & 98.01 \\
                      & MB (2.0 bpp)          & 100.00 & 100.00 & 98.18 \\
                      & MB (3.0 bpp)          & 100.00 & 100.00 & 99.14 \\
                      & MC (1.0 bpp)          & 99.98 & 99.97 & 92.72 \\
                      & MC (2.0 bpp)          & 99.98 & 99.97 & 90.99 \\
                      & MC (3.0 bpp)          & 99.99 & 99.98 & 93.48 \\
                      & Multi-bits (4.0 bpp)  & 99.98 & 99.98 & 96.34 \\
                      & Multi-bits (5.0 bpp)  & 99.94 & 99.93 & 93.16 \\
                      & Multi-bits (6.0 bpp)  & 99.91 & 99.89 & 91.58 \\
                      & Multi-bits (9.0 bpp)  & 99.20 & 99.13 & 83.18 \\
                      \midrule
\multirow{13}{*}{FFHQ} & MN (1.0 bpp)          & 98.60 & 98.45 & 87.92 \\
                      & MN (2.0 bpp)          & 98.56 & 98.38 & 86.64 \\
                      & MN (3.0 bpp)          & 98.60 & 98.42 & 87.00 \\
                      & MB (1.0 bpp)          & 100.00 & 100.00 & 97.96 \\
                      & MB (2.0 bpp)          & 100.00 & 100.00 & 98.10 \\
                      & MB (3.0 bpp)          & 100.00 & 100.00 & 98,64 \\
                      & MC (1.0 bpp)          & 100.00 & 99.99 & 93.33 \\
                      & MC (2.0 bpp)          & 100.00 & 99.99 & 92.16 \\
                      & MC (3.0 bpp)          & 100.00 & 100.00 & 93.46 \\
                      & Multi-bits (4.0 bpp)  & 100.00 & 99.99 & 95.61 \\
                      & Multi-bits (5.0 bpp)  & 99.99 & 99.97 & 92.45 \\
                      & Multi-bits (6.0 bpp)  & 99.99 & 99.97 & 91.08 \\
                      & Multi-bits (9.0 bpp)  & 99.71 & 99.64 & 82.25 \\
                      \midrule
\multirow{13}{*}{CIFAR-10} & MN (1.0 bpp)          
                                              & 97.55 & 97.38 & 84.51 \\
                      & MN (2.0 bpp)          & 97.54 & 97.36 & 83.72 \\
                      & MN (3.0 bpp)          & 97.62 & 97.36 & 84.49 \\
                      & MB (1.0 bpp)          & 100.00 & 100.00 & 94.51 \\
                      & MB (2.0 bpp)          & 100.00 & 99.99 & 94.23 \\
                      & MB (3.0 bpp)          & 100.00 & 99.99 & 94.78 \\
                      & MC (1.0 bpp)          & 99.97 & 99.94 & 84.91 \\
                      & MC (2.0 bpp)          & 99.97 & 99.93 & 84.54 \\
                      & MC (3.0 bpp)          & 99.98 & 99.95 & 85.26 \\
                      & Multi-bits (4.0 bpp)  & 99.94 & 99.91 & 89.54 \\
                      & Multi-bits (5.0 bpp)  & 99.90 & 99.85 & 86.16 \\
                      & Multi-bits (6.0 bpp)  & 99.86 & 99.79 & 83.98 \\
                      & Multi-bits (9.0 bpp)  & 98.93 & 98.71 & 75.96 \\ \bottomrule
\end{tabular}
\label{tiff_table}
\end{table}

\section{Implementation Details}

In our experiments, we use the official code of PFGM++~\cite{xu2023pfgm++} built upon the official code of EDM~\cite{karras2022elucidating}.
We adopt the same hyper-parameters as EDM and PFGM++.
We set $\sigma_{\mathrm{max}} = \sigma_T = 80, \sigma_{\mathrm{min}} = 0.002$, and $\rho = 7$. 
These hyper-parameters determine the standard deviation of $x(t)$, which represents the noise in the denoising process.
When the number of denoising inference steps is $N$, the standard deviation of the noise after denoising $i$ times is as follows:
\begin{align}
     {\sigma_{\mathrm{max}}}^{\frac{1}{\rho}} + \frac{i}{N-1} ({\sigma_{\mathrm{min}}}^{\frac{1}{\rho}}  - {\sigma_{\mathrm{max}}}^{\frac{1}{\rho}}).   
\end{align}
For our method, we use Heun's sampler~\cite{karras2022elucidating}, which is based on Heun's method~\cite{ascher1998computer} for sampling images.
Heun's method is a numerical integrator that reduces errors through performing two calculations in each step.
In the extraction function, denoted as $f_\theta^{-1}$, we employ Heun's method to estimate $\mathbf{z}_m$, which corresponds to the reversible calculation of Heun's sampler using the same hyper-parameters.

\section{Potential Social Impact}
\label{appendix_impact}

We propose a powerful generative steganography that can enhance information security. 
In scenarios involving wiretapping and vulnerable communications, \our{} provides protection against third-party players attempting to access user information. 
However, it also introduces certain risks as malicious individuals or industrial spies may exploit it to compromise corporate confidentiality. 
Since our generative steganography approach relies on pre-trained diffusion models, it becomes susceptible to such exploitation. 
Further research in steganalysis and developing safeguards against the misuse of pre-trained diffusion models is necessary to address these risks.

\section{Additional Samples}

We present additional samples of stego images generated by pre-trained models trained on AFHQv2 and FFHQ 64$\times$64 datasets in Figure~\ref{app_afhq}, \ref{app_ffhq}, and \ref{app_high64}. 
The stego images generated using the CIFAR-10 dataset are displayed in Figure~\ref{app_cif_low} and \ref{app_cif_high}.
\newpage
{
\begin{figure}[t]
    \centering
    \includegraphics[width = 0.7\textwidth]{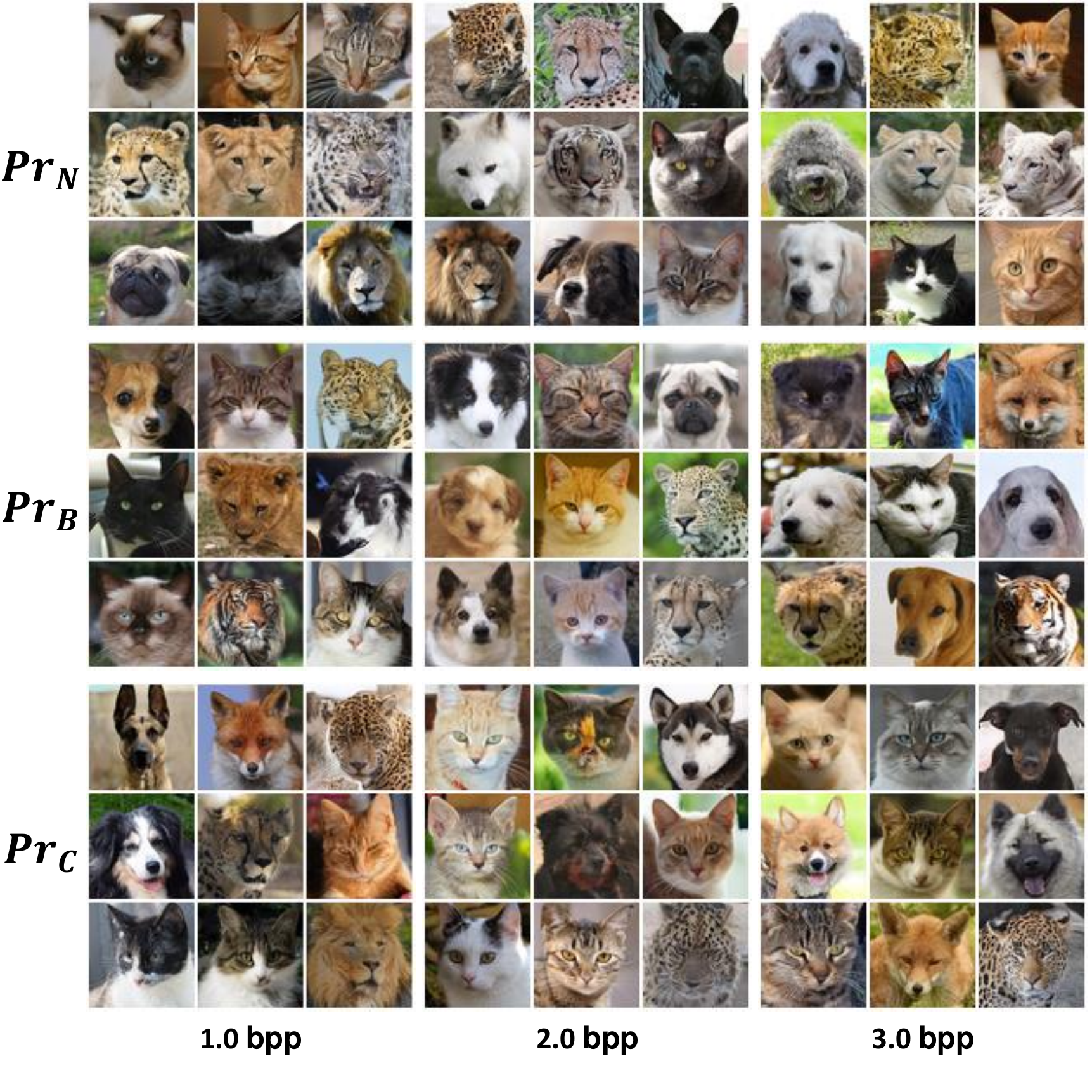}
    \caption{AFHQv2 64×64 stego images with pre-trained EDM models.}
    \label{app_afhq}
\end{figure}
\begin{figure}[b]
    \centering
    \includegraphics[width = 0.7\textwidth]{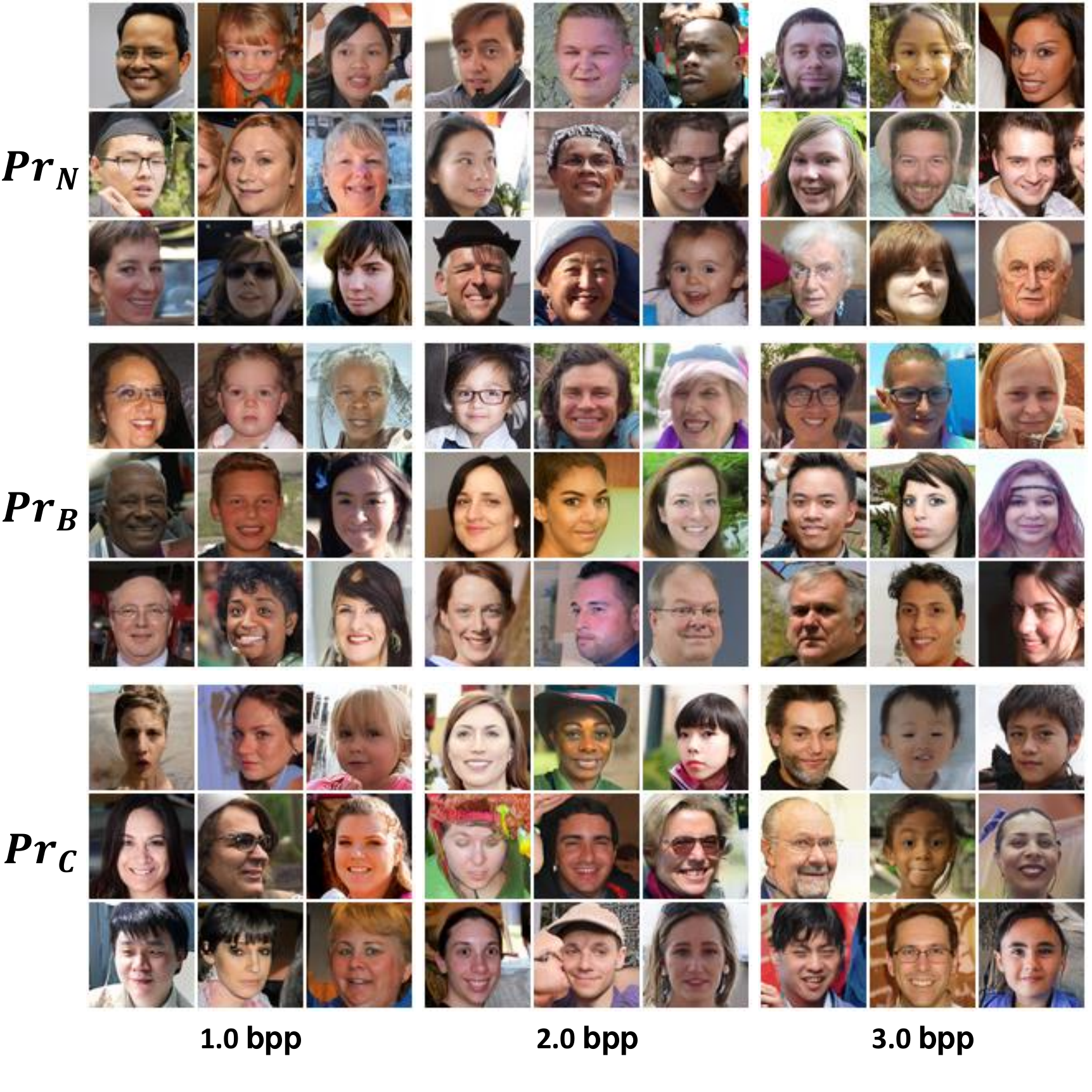}
    \caption{FFHQ 64×64 stego images with pre-trained EDM models.}
    \label{app_ffhq}
\end{figure}
\begin{figure}[tbh]
    \centering
    \includegraphics[width = 0.7\textwidth]{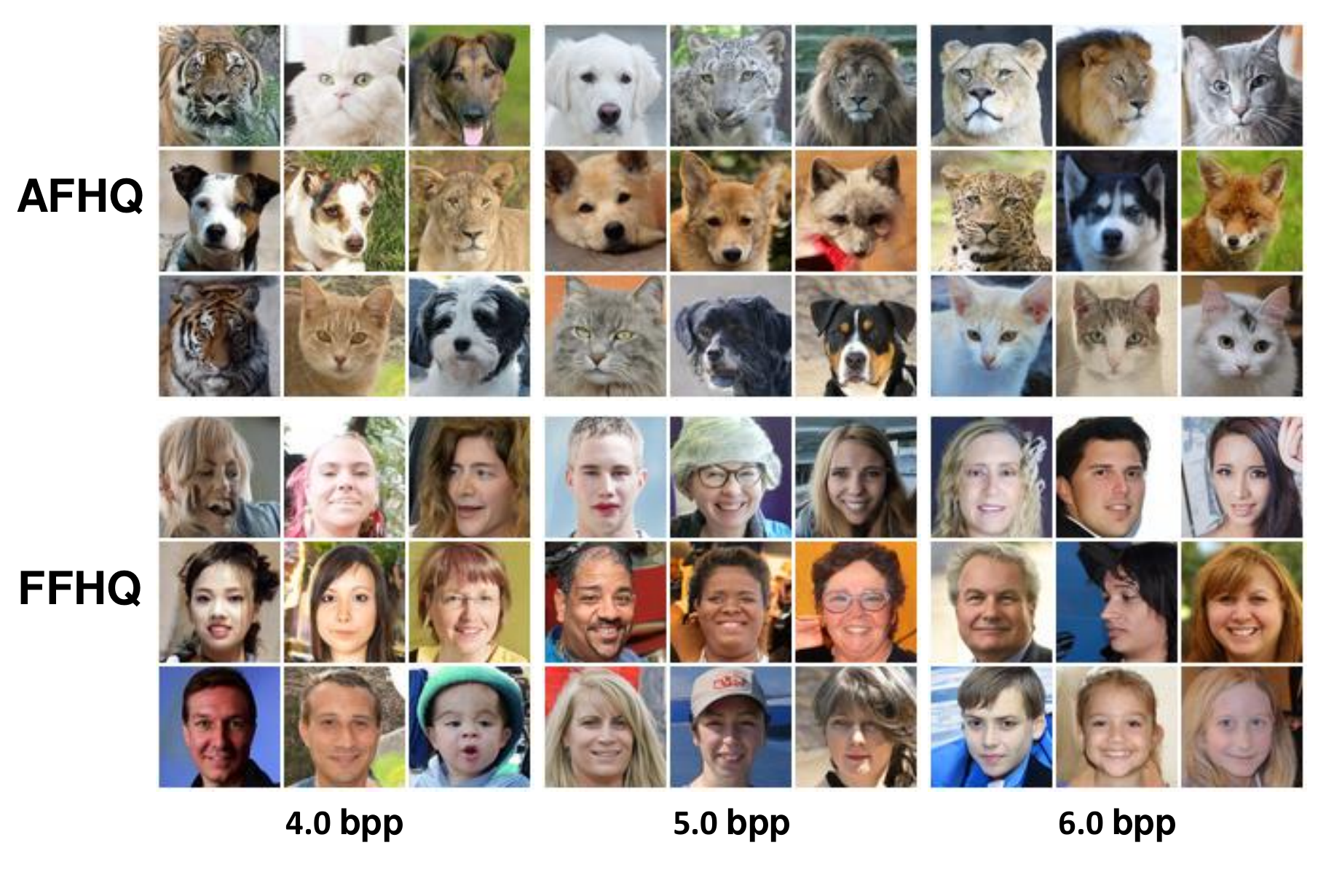}
    \caption{Stego images with pre-trained EDM models and the multi-bits projection.}
    \label{app_high64}
\end{figure}
\begin{figure}[tbh]
\centering
    \subfigure[MN]{\includegraphics[width=0.3\textwidth]{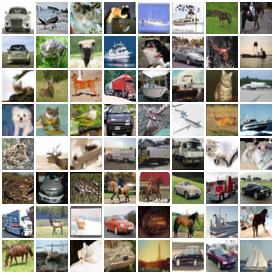}}
    \subfigure[MB]{\includegraphics[width=0.3\textwidth]{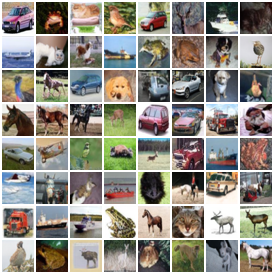}}
    \subfigure[MC]{\includegraphics[width=0.3\textwidth]{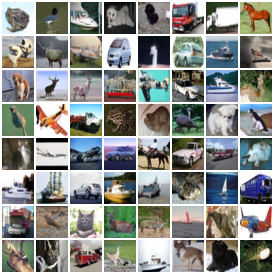}}
    \caption{CIFAR-10 stego images with pre-trained EDM models hiding 3.0 bpp messages.}
    \label{app_cif_low}
\end{figure}
\begin{figure}[tbh]
\centering
    \subfigure[4.0 bpp]{\includegraphics[width=0.3\textwidth]{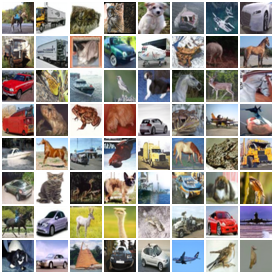}}
    \subfigure[5.0 bpp]{\includegraphics[width=0.3\textwidth]{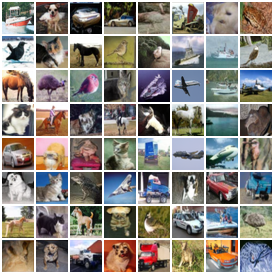}}
    \subfigure[6.0 bpp]{\includegraphics[width=0.3\textwidth]{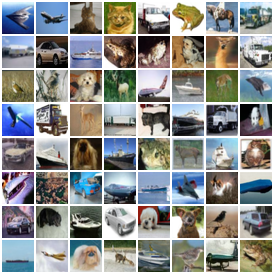}}
    \caption{CIFAR-10 stego images with pre-trained EDM models and the multi-bits projection.}
    \label{app_cif_high}
\end{figure}
}

\end{document}